\documentclass[conference,letterpaper]{IEEEtran}
\IEEEoverridecommandlockouts

\usepackage{cite}
\usepackage{amsmath,amssymb,amsfonts}
\usepackage{algorithmic}
\usepackage{graphicx}
\usepackage{textcomp}
\usepackage{xcolor}
\usepackage{color}														
\usepackage{colortbl}
\usepackage{siunitx}
\usepackage[english]{babel}	
\usepackage{multirow}
\usepackage{hyperref}
\addto\captionsenglish{\renewcommand{\figurename}{Fig.}}
\def\BibTeX{{\rm B\kern-.05em{\sc i\kern-.025em b}\kern-.08em
    T\kern-.1667em\lower.7ex\hbox{E}\kern-.125emX}}
\newif\ifcopyright
	\copyrighttrue

\begin{document}
	\ifcopyright
	{\LARGE IEEE Copyright Notice}
	\newline
	\fboxrule=0.4pt \fboxsep=3pt
	
	\fbox{\begin{minipage}{1.1\linewidth}  
			Copyright (c) 2021 IEEE. Personal use of this material is permitted. For any other purposes, permission must be obtained from the IEEE by emailing pubs-permissions@ieee.org. \\
			
			Accepted to be published in: Proceedings of the 2021 IEEE/ASME International Conference on Advanced Intelligent Mechatronics (AIM), July 12 – 16, 2021, Delft, The Netherlands.  
			
	\end{minipage}}
	\else
	\fi
	
\hyphenation{LiDAR}	
\graphicspath{{./Abbildungen/}}
\title{Have I been here before? Learning to Close the Loop \mbox{with LiDAR Data in Graph-Based SLAM}
\thanks{$^{1}$T.-L. Habich, M. Stuede and S. Spindeldreier are with the Leibniz University Hannover, Institute of Mechatronic Systems, D-30823 \mbox{Garbsen, Germany,}
	{tim-lukas.habich@imes.uni-hannover.de}

	$^{2}$M. Labbé is with the Université de Sherbrooke, Interdisciplinary Institute of Technological Innovation, Sherbrooke, Québec, J1K 0A5 Canada
	}

\author{\IEEEauthorblockN{Tim-Lukas Habich$^{1}$, Marvin Stuede$^{1}$, Mathieu Labbé$^{2}$ and Svenja Spindeldreier$^{1}$}
	
	}}%
\maketitle
\begin{abstract}
This work presents an extension of graph-based SLAM methods to exploit the potential of 3D laser scans for loop detection. 
Every high-dimensional point cloud is replaced by a compact global descriptor, whereby a trained detector decides whether a loop exists. 
Searching for loops is performed locally in a variable space to consider the odometry drift. 
Since closing a wrong loop has fatal consequences, an extensive verification is performed before acceptance. 
The proposed algorithm is implemented as an extension of the widely used state-of-the-art library RTAB-Map, and several experiments show the improvement:
During SLAM with a mobile service robot in changing indoor and outdoor campus environments, our approach improves RTAB-Map regarding total number of closed loops.
Especially in the presence of significant environmental changes, which typically lead to failure, localization becomes possible by our extension.
Experiments with a car in traffic (KITTI benchmark) show the general applicability of our approach. 
These results are comparable to the state-of-the-art LiDAR method LOAM. 
The developed ROS package is freely available.\\ 
\end{abstract}
\section{Introduction}
One key challenge in developing autonomous mobile robots is the SLAM (Simultaneous Localization And Mapping) problem \cite{Thrun2008}. 
Graph-based solutions are often based on a static environment, which is unrealistic: objects move, seasons influence the appearance of the surroundings and -- depending on the lighting -- images of the same scene differ. 
For long-term autonomy in dynamic environments two main requirements must be met. 
First, the computational complexity of graph optimization, the back-end, must be limited. 
Second, the process has to be robust against changes in the environment. 
The processing of sensor data, the front-end, must be developed in such a way that, despite changes, already visited places are recognized again. 
Answering the question \textit{Have I been here before?} belongs to the most important tasks during SLAM, since in presence of a loop closure, the odometry drift can be corrected retrospectively and the map quality can be improved.

\inkscapescale{nodes_detector}{Loops with discrete nodes (blue points) are searched for in a variable radius of the current position (orange arrow). 
	The search space depends on the position uncertainty.
	Due to the use of different memories not all nodes in the radius are used as loop candidates.}{htbp}{.85}

Loop detection approaches can be classified according to the sensor technology used. 
Due to the high information density in an image and the multitude of effective techniques, visual methods are widely used. 
Where such methods are stable for static environments, the illumination change can already lead to failure within a few hours. 
Actively illuminated sensors such as LiDAR (Light Detection And Ranging) scanners are robust in such situations, as generated point clouds are illumination-invariant. 
This work deals with the combination of the two approaches to realize robust mapping and localization. 
The basis for this is the widespread library RTAB-Map (Real-Time Appearance-Based Mapping) \cite{Labbe.2018b}. 
Although the method can be used with a LiDAR, the data is only marginally utilized due to the simplified method of loop search within a constant radius using 2D Iterative Closest Point (ICP), making an extension suitable. 
Fig.~\ref{fig:nodes_detector} illustrates our approach. 
Loops are searched based on scan descriptors, and within a variable radius $r$ depending on the largest eigenvalue $\lambda_{\textrm{max}}$ of the position covariance matrix to take the odometry drift into account. 
If a loop exists, the respective scans are registered and the graph is optimized with the calculated transformation $\mm{T}$.

The paper is structured as follows. Sec.~\ref{relatedwork} presents related work.
Based on this, Sec.~\ref{cmrlidarloop} describes our contributions:
\begin{itemize}
	\item further development of a loop detection method to enable the necessary scan registration,
	\item introduction of several loop verification steps to reject false positives,
	\item elaboration of a robust, open-source ROS package\footnote{\url{https://github.com/MarvinStuede/cmr\_lidarloop}} to close loops with LiDAR data during graph-based SLAM.
\end{itemize}
This is followed by various validations in Sec.~\ref{expval} and conclusions in Sec.~\ref{conclusions}.

\section{Related Work}\label{relatedwork}
Approaches to mapping changing environments are manifold. 
Examples are filtering moving objects and assuming a static environment \cite{Hua.2016}, modeling the dynamics in the frequency domain \cite{Krajnik.} and integrating new data to adapt the map \cite{Krajnik.2016}. 
The continuous adaptation to changes is also possible by removing old nodes at loop closure \cite{Lazaro.01.10.201805.10.2018} or by representing the environment in several time frames simultaneously \cite{Biber.June8112005}. 

RTAB-Map \cite{Labbe.2018b} is suitable for large-scale and long-term online SLAM in changing environments due to its memory management \cite{Labbe.2018}:
The bounded Working Memory (WM) ensures a bounded demand on computing which is realized by transferring nodes to a database. 
A buffer of constant size ensures that recently created nodes are not considered for loop detection, since the odometry drift cannot be significantly corrected. 
Whereas most state-of-the-art methods are either visual or LiDAR-based, RTAB-Map supports both. 
This key advantage enables the use of many sensor configurations, comparison of results and easy integration into different systems.
The availability in the ROS framework, multi-session operation and a high accuracy \cite{Labbe.2018b} further promote its deployment.

RTAB-Map's LiDAR-based \textit{proximity detection} module was developed for challenging situations, such as a significant illumination change, in which the visual loop detection is not promising.
However, the used 2D ICP algorithm is a simplified method for loop detection.
The high amount of data within the three-dimensional point cloud is not used and a fast loop search with several hundred potential pairs is not possible. 
Beyond that, this module only searches within a constant radius for loops which is problematic, since the error of the estimated position increases due to odometry drift.
The search space should thus be permanently adjusted. 
Further, a local search is useless at the beginning of multi-session operation, because the position in the old map is unknown.
Instead, a global search must be performed here so that a link can be established between the two sessions, and the robot can locate itself in the old map.  
The potential of scans to close loops is thus not exploited, making an extension necessary.

To detect loops with LiDAR data, there are a variety of methods which are based on histograms to reduce the scan dimension. 
Representing a scanning area as a piecewise continuous function using Normal Distribution Transform (NDT) and detecting loops by matching feature histograms realizes a high accuracy \cite{Magnusson.2009}. 
The time to create such histograms is an important property which is a disadvantage of NDT. 
Instead, by using simpler histrogram types such as the range or height of each point, a fast histogram generation \cite{Rohling.2015} is possible with results comparable to NDT.
The azimuthal and radial division of the scan into bins and use of the maximum height of the points in each bin also allows a fast computation of a global descriptor called Scan Context \cite{gkim-2018-iros}. 
However, besides height and range, there is further information in the raw scan, which can be obtained with low computational effort.
In \cite{Granstrom.}, each point cloud is described with 41 rotationally invariant geometric features, so that a trained classifier decides whether a loop exists. 
The extensive scan description with small, rapidly computed features and fast prediction make this approach suitable for efficient loop search.
However, the method does neither consider the scan registration nor the important loop verification to reject false positives.
We thus extend the method of Granström and Schön \cite{Granstrom.} with these essential elements and integrate it into RTAB-Map, so that a robust loop search with laser scans is realized.

\section{SLAM with Learned LiDAR Loop Detector}\label{cmrlidarloop}
\begin{figure*}[h]
	\centering
	\resizebox{1\linewidth}{!}{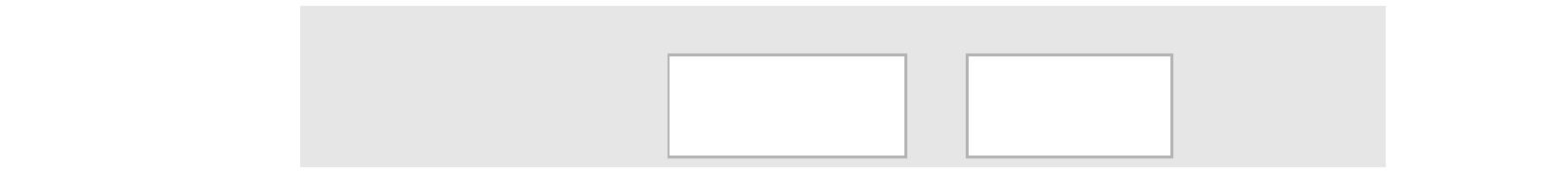}
	\caption{LiDAR extension of RTAB-Map. Loops are permanently searched for using the scan descriptors. In case of a detected loop, the registration of the scans takes place in a parallel thread. After the point clouds have been pre-processed, first a rough registration and then a refinement takes place.} \label{fig:cmr_lidarloop}
\end{figure*}

The developed LiDAR extension is presented in Fig.~\ref{fig:cmr_lidarloop} and described in this section.
\subsection{Loop Classification}\label{lidarclass}
Loop detection is a binary problem: either the robot has already visited the current location in the past or not. 
To solve this problem based on LiDAR data, a classifier is trained. 
A point cloud $\m{P}{}{}{}{\mathit{i}}=\{\m{p}{}{}{\mathit{k}}{\mathit{i}}\}^{N}_{\mathit{k}=1}$ from node $i$ in the graph contains $N$ points $\m{p}{}{}{\mathit{k}}{\mathit{i}}\in\mathbb{R}^3$ representing the environment. 
The rotation invariant classifier presented in \cite{Granstrom.} is used in our work and is briefly introduced in the following.
 
Depending on the LiDAR, the number of points differs. 
For example, the scanner of the mobile robot Sobi used for evaluation (see Sec.~\ref{expval}) generates $N$$\approx\,$45,000 points per cloud, requiring a dimensionality reduction for comparison. 
For this, each cloud is described by global features \cite{Granstrom.}
\e{\m{f}{}{}{}{\mathit{i}}=(\m{f}{}{}{I}{\mathit{i}},\m{f}{}{}{II}{\mathit{i}})^\textrm{T}.}{feature_vector}
Features $\m{f}{}{}{I}{\mathit{i}}$ of type I map the point cloud to a real number, e.g. simple geometric quantities such as the mean range or the range's standard deviation. 
More complex geometric quantities, such as the center and radius of a sphere fitted to the point cloud, are also calculated. 
A total of 32 features of type I are computed. 
In addition, there are nine features $\m{f}{}{}{II}{\mathit{i}}$ of type II.
These are range histograms with nine different container sizes $b_{1},\ldots,b_{9}$. 
For each histogram, starting at the sensor origin, the scan is divided into annular container of constant width and the points lying in each container are counted. 
Each feature is a vector whose dimension depends on the container size.
Before feature computation, every scan is processed in a way that all points with range $r_k$ greater than a maximum range $r_\textrm{max}$ are moved in the direction of the sensor origin so that $r_k \leq r_\textrm{max}$ applies for all points. 
This limitation allows range histograms of one container size to have the same dimension in any case. 
A total of 41 features are calculated consisting of 843 real numbers during SLAM with Sobi (see Sec.~\ref{expval}). 
Thus, the dimension of any scan is significantly reduced.

If during mapping and localization the features are calculated and stored for each node, the descriptor $\m{f}{}{}{}{c}$ of the current position can be efficiently compared to descriptors of map nodes $\m{f}{}{}{}{\mathit{i}}$ with respect to a possible loop closure. 
An AdaBoost Classifier \cite{Pedregosa.2011} \cite{Hastie.2009} predicts if a loop is present and the entity $\sk{p}{}{\mathit{i}}{}{c}$ indicates the probability that the corresponding scans are from the same location. 
The input of the classifier is generated from the respective feature vectors by comparing them appropriately \cite{Granstrom.}.
Type I features are compared via the absolute difference and histogram comparison for features of type II is done using Pearson's correlation coefficient.

AdaBoost classification \cite{Freund.1997} belongs to the class of boosting algorithms and uses $T$ weak learners, which together form a strong classifier. 
In each learning round, the data is reweighted depending on the current prediction error. 
Incorrectly classified and thus difficult cases are prioritized higher, so that the number of training rounds $T$ significantly influences the accuracy. 
For $T\geq50$ weak classifiers, the error was observed to stop decreasing for the used LiDAR descriptors \cite{Granstrom.}. 
Accordingly, in the present work $T=50$ rounds are performed for learning.

\subsection{Scan Registration}\label{lidarreg}
In case of a detection, a further step is necessary to close a loop: the scan registration. 
This aims at determining the homogeneous transformation $\mm{T}$ between the current point cloud $\m{P}{}{}{}{c}$ and the point cloud $\m{P}{}{}{}{*}$, with which a loop was detected. 
The transformation is then used to add a link between the two nodes in the map in order to take the loop into account.
The entire registration process is illustrated in Fig.~\ref{fig:cmr_lidarloop} and is implemented with the point cloud library \cite{Rusu.09.05.201113.05.2011}. 

Registering raw point clouds is computationally expensive and the scans also contain regions that prevent successful registration. 
Examples are points on the ground or regions generated by reflection of the laser beam on glass fronts.
To encounter this, the following filters are executed consecutively:
\begin{itemize}
	\item voxel grid filter with voxel side length $l$,
	\item height filter to remove points with height lower than $z_\textrm{lim}$,
	\item intensity filter to remove points with intensity lower than $i_\textrm{lim}$ and therefore points which laser beams were reflected,
	\item range filter to remove points with range larger than $r_\textrm{lim}$ and therefore to remove scan regions which are far away and do not describe the environment in sufficient detail,
	\item random downsampling filter to randomly remove points until the point cloud consists of $n_\textrm{p,max}$ points.
\end{itemize}

The pre-processed scans $\m{\tilde{P}}{}{}{}{c}$, $\m{\tilde{P}}{}{}{}{*}$ used for registration can have a large translational and rotational offset. 
Using a local method like ICP \cite{Besl.1992} would not be expedient due to a convergence into a local optimum. 
Consequently, a rough registration to compute an initial alignment takes place first according to \cite{Holz.2015}. 
Thereby, Fast Point Feature Histograms (FPFH) \cite{Rusu.2009} are used as robust multidimensional features that describe local geometry around a point and only persistent features are used \cite{Rusu.2008}. 
The latter increases robustness, since only unique regions are taken into account. 
With these keypoints, the correspondences can then be estimated by performing a nearest neighbor search in the feature space.

Because of a partial overlap, not every keypoint has a match and a large number of incorrect correspondences would be possible. 
These outliers negatively influence the registration and must be rejected. By means of outlier rejection based on RANdom SAmple Consensus (RANSAC) \cite{Fischler.1981} a transformation is computed over a multiplicity of iterations with different subsets of correspondences. 
Only those correspondences are classified as correct whose Euclidean distance, after application of the respective transformation, is smaller than a defined threshold. 
The outliers that are present during the transformation with the most inliers are rejected.

With the filtered correspondences the initial alignment can be completed using Singular Value Decomposition (SVD) \cite{Horn.1987}. 
The transformation $\m{T}{}{}{\,\textrm{IA}}{}$ of this global method is the initial guess for subsequent refinement, so that the output of the ICP registration $\mm{T}$ is used to correct the odometry drift. 
Since adding a wrong transformation has drastic consequences for the SLAM procedure, some verification criteria must be met. 
First, both processed scans must consist of at least $n_\textrm{p,min}$ points so that enough data is included. 
Second, a transformation is only accepted if at least $n_\textrm{inliers}$ correspondences after outlier rejection are present which ensures that enough matching areas are found. 
Third, the transformation has a translational offset of at most $t_\textrm{max}$. 
This condition increases the robustness since a larger translational offset leads to a smaller scan overlap, and thus the registration can become less accurate. 

\subsection{Extending the RTAB-Map Framework}\label{impl}
The presented loop detection and scan registration are integrated as an extension into RTAB-Map which is illustrated in Fig.~\ref{fig:cmr_lidarloop}. 
For each scan, the corresponding descriptor is calculated and sent to RTAB-Map.  
Saving the scans with associated features enables compatibility with multi-session operation.  
The extension is continuously supplied with current map data which contains required information of the graph such as the position $\m{x}{}{}{}{c}$ of the current node. 
Further, the extension receives the positions $\mm{X}=\{\m{x}{}{}{}{\mathit{i}}|\forall i\}$ of all map nodes and associated descriptors $\mm{F}=\{\m{f}{}{}{}{\mathit{i}}|\forall i\}$, which is the basis for the actual task to detect loops.
 
When receiving new map data, a loop search is done with the trained detector. 
Similar to \textit{proximity detection}, the current node is compared with other WM nodes within a certain radius only. 
For example, if the robot is located in a corridor and the entire WM is checked, a loop detection with a node in a similar corridor at a completely different location in the building would be possible due to similar scans. 
Adding such a loop would be fatal for the SLAM process, so the local restriction is useful for increasing robustness.

In determining the search space, the present work differs from RTAB-Map's approach. 
For large loops, a large odometry drift is present and therefore, a large uncertainty in the estimated position. 
The search carried out in \textit{proximity detection} in a constant radius is problematic here because the estimated position is highly inaccurate and loops are searched for in the wrong map area. 
Accordingly, we follow a different philosophy: the use of a variable search space depending on the position inaccuracy. 
The search radius 
\e{r(\lambda_{\textrm{max}})=r_\textrm{min}+\beta g_{\textrm{max}}(\lambda_{\textrm{max}})}{r_search}
consists of two parts. 
In the constant radius $r_\textrm{min}$, loops are searched for if the current position can be assumed to be without errors. 
This is the case at the start of the process as well as at loop closures, where the inaccuracy is corrected. 
In all other cases, the estimated position is subject to error, whereby we assume the odometry error to be steadily increasing. 
This property is taken into account by the second component of (\ref{eq:r_search}). 
Thereby, $g_{\textrm{max}}(\lambda_{\textrm{max}})=2\sqrt{5.991\lambda_{\textrm{max}}}$ is the length of the longest major axis of the $95$\% confidence ellipse of the two-dimensional position. 
This quantity can be obtained from the largest eigenvalue $\lambda_{\textrm{max}}$ of the covariance matrix, for which the odometry data is processed by the extension. 
The parameter $\beta$ serves as a scaling factor. 
Fig.~\ref{fig:R_search} visualizes the variable search space. 
The approach enables to include nodes close to the exact position in the loop search despite a large position inaccuracy.

\inkscapescale{R_search}{Despite the difference between the actual (green arrow) and estimated position (orange arrow), the search space (orange circle) contains the desired circular area with the radius $r_\textrm{min}$ around the actual position. 
	The schematic course shows the increase of the radius with increasing time $t$ due to odometry error. 
	If a loop closure occurs, the uncertainty is reset.}{bp}{1}

As already mentioned, adding a wrong loop is fatal for the integrity of the map. 
Hence, further precautions are taken to reduce the number of false positives. 
Only when $\sk{p}{}{\mathit{i}}{}{c}>\sk{p}{}{}{}{min}$ applies to the predicted loop probability $\sk{p}{}{\mathit{i}}{}{c}$, the pair is treated as a potential loop pair. 
Further information on the adjustment of the threshold $\sk{p}{}{}{}{min}$, i.e. the fine tuning of precision and recall, is presented in Sec.~\ref{det_perf}.

If the search was successful, a final verification is performed with the node of highest loop probability $\sk{p}{}{*}{}{c}$ at the location $i^*$. 
The current node is compared with the surrounding nodes of the potential loop candidate. 
If at least one loop with the current node is detected in the immediate neighborhood consisting of $2n_\textrm{v}$ verification nodes, this candidate is accepted. 
In each case $n_\textrm{v}$ nodes with $i<i^*$ and $n_\textrm{v}$ nodes with $i>i^*$ are considered. 
For the strictest variant ($n_\textrm{v}=1$) there must be either another loop with the node created before or after the loop candidate. 
Otherwise, the loop is rejected and the search starts again when new map data is received. 
If the verification is successful, the scan registration takes place in a parallel thread and the link between the involved nodes is added with the accepted transformation $\mm{T}$. 
After further verification in RTAB-Map, the loop can be closed by graph optimization. 

Two final adjustments remain to be made: 
First, loops must be searched robustly even in multi-session operation, which is impossible with the previous approach of a local search if the position is unknown. 
Especially at the beginning of a new session the relative position in the old map is in general not given -- the initial state problem. 
A position-independent search must be carried out here, so that nodes of the entire WM are examined. 
As alternative local restriction during multi-session, the size $n_\textrm{ms}$ is introduced which is the number of consecutive loop candidates that must lie within a radius $r_\textrm{ms}$. 
Only one loop is accepted if several loop pairs are from the same place. 
This type of loop search is done in multi-session mode for the first $n_\textrm{start}$ accepted loops, after which it has to be checked if the localization in the old map was successful. 
The criterion for this is the ratio $\alpha=\frac{n_\textrm{local}}{n_\textrm{WM}},$ where $n_\textrm{local}$ is the number of nodes in the local map and $n_\textrm{WM}$ is the number of all WM nodes. 
If this ratio is smaller than a defined threshold $\alpha_\textrm{min}$, there are few nodes in the local map, so the relative positions of many WM nodes to the robot are unknown. 
In this case, the entire WM should continue to be searched for loops, since localization is not yet satisfactory. 
Otherwise, the relative position of many WM nodes with regard to the robot is known and the local search in the variable radius can be started. 
To meet online requirements, the second adjustment is to limit the number of nodes used in the search. 
For this, $n_\textrm{n,max}$ nodes are randomly selected from the possible candidates. 
Due to the fast detector, hundreds of nodes can be used. 

\section{Experimental Validation}\label{expval}
The evaluation is divided into three parts:
First, a loop detector is trained and tested in an independent environment (see Sec.~\ref{det_perf}). 
Second, multi-session experiments were performed with this detector under challenging conditions (see Sec.~\ref{val_slam}). 
The mobile service robot Sobi \cite{Ehlers.2020} is used for both validations, which is a ROS-based information and guidance system equipped with a differential drive base (Neobotix MP-500), two RGBD cameras (Intel Realsense D435, front and back) and a 3D LiDAR (Velodyne VLP-16). 
An extended Kalman filter \cite{Moore.2016} is used to fuse the wheel odometry with the IMU data (XSens MTi-30).
Third, the general applicability of our approach is shown with the widely used KITTI\cite{Geiger.2012b} dataset in Sec.~\ref{kitti_sec}. 

\subsection{Detector Performance}\label{det_perf}
To train the detector, indoor and outdoor data consisting of descriptors with corresponding coordinates was collected on an university campus. 
There are large distances between objects in outdoor areas of the campus, so $r_\textrm{max}=40\textrm{m}$ is chosen. 
With the container sizes $b_{1},\ldots,b_{9}$ suggested in \cite{Granstrom.}, each feature vector consists of 843 entries. 
There is no ground truth position, so the optimized poses were obtained from RTAB-Map to generate the most accurate positions of the 1248 nodes with a path length of 697m.

Comparing each node with itself and with all others gives 779,376 pairs. 
As the distance at which the detector should treat a pair as positive, we choose 3m. 
With this distinction, the set is divided into 11,458 positive and 767,918 negative pairs. 
Training with this set would not be expedient, as the number of pairs for the classes is clearly unbalanced. 
As in \cite{Granstrom.}, a random subset of negative pairs is used, where we select 11,458 pairs to obtain the same amount of data for both classes. 
However, it is noticeable that the performance varies considerably depending on the subset, since half of the data is randomly taken from a large set. 
Thus, an optimization was carried out, whereby 50 detectors were trained with different subsets. 
For comparison, common criteria consisting of detection rate $D$ and false alarm rate $FA$ are used: 
\e{D=\frac{\textrm{\# Positive data pairs classified as positive}}{\textrm{\# Positive data pairs}},}{D}
\e{FA=\frac{\textrm{\# Negative data pairs classified as positive}}{\textrm{\# Negative data pairs}}.}{FA} 

\begin{figure*}[t]
	\centerline{\includegraphics[width=\linewidth]{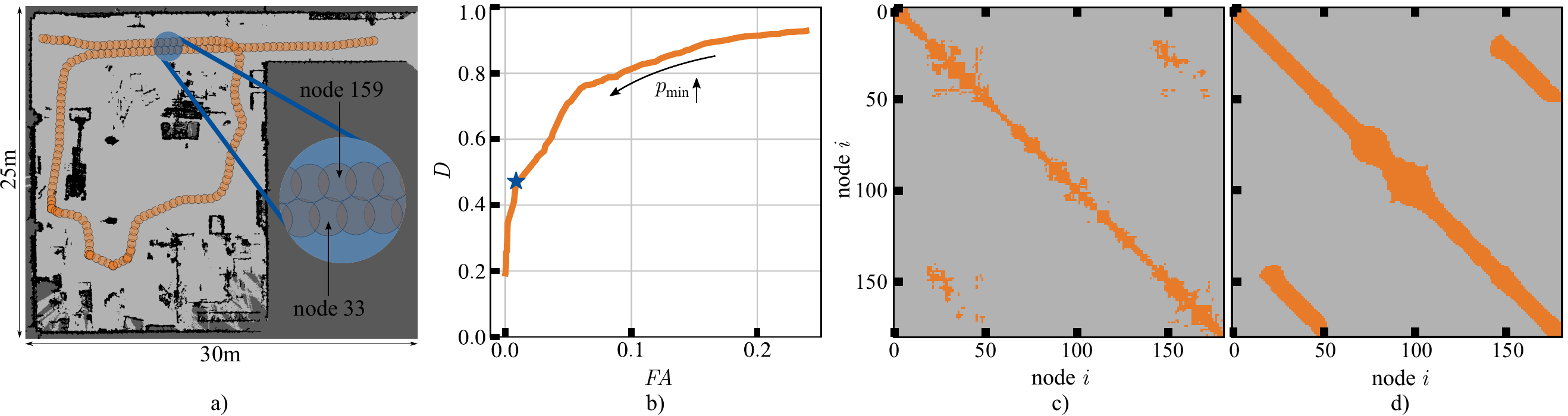}}
	\caption{Detector performance for training-independent test environment a). ROC b) shows performances of best detector at different thresholds $p_\textrm{min}$. Detectors can be compared using their detection rates at the desired boundary of $FA<1\%$ (blue star). Corresponding classification matrix c) compared to distance matrix d) visualizes performance of this detector at $p_\textrm{min}=52.4\%$ with $D=47.3\%$ and $FA=0.8\%$. Orange areas represent loops and grey areas no loops.
	}
	\label{fig:detector_performance}
\end{figure*}

An ideal detector would detect every loop ($D=100\%$) and would also not detect any loop incorrectly ($FA=0\%$). 
Since our extension still verifies every possible loop pair, $FA<1\%$ is set as target. 
The decisive parameter for this is the introduced threshold $p_\textrm{min}$, which is fine tuned with a training-independent test dataset consisting of 181 nodes.
The path of length 72m is a loop in an indoor hall, which is illustrated in Fig.~\ref{fig:detector_performance}a). 
For each of the 50 detectors, $p_\textrm{min}$ was incrementally increased so that the requirement $FA<1\%$ is met.
This process can be visualized with the ROC (Robot Operating Curve) of the best performing detector in Fig.~\ref{fig:detector_performance}b).
Increasing the threshold results in a more restrictive detector, which is less erroneous but also detects fewer loops.
The best classifier ($D=47.3\%,\,FA=0.8\%,\,p_\textrm{min}=52.4\%$) detects approximately half of all loops with only few false positives in the unknown environment.

The test results can be illustrated in the form of two matrices. 
The classification matrix Fig.~\ref{fig:detector_performance}c), in which each node pair is examined by the detector for a loop, is compared with the distance matrix  Fig.~\ref{fig:detector_performance}d). 
The latter represents the underlying ground truth, since all pairs with a distance less than 3m are treated as loop and the rest as negative pairs. 
The orange area on the diagonal represents the successive nodes that are close enough together. 
The off-diagonal orange area is the loop that is driven, since the robot returned to an old position at a later point in time and the nodes concerned are a short distance apart. 
For the trained detector, these matrices are roughly similar and all detected loops are located in areas where the pairs are not far away from each other. 
However, pairs with a distance minimally greater than the selected 3m are classified as positive as well. 
This does not pose a problem, as these loop closures are also desirable.

\subsection{SLAM under Challenging Conditions}\label{val_slam}
The presented method is furthermore validated in an experiment under challenging conditions. 
Since the introduced extension shall support the loop detection, we compare RTAB-Map extended with our work against the default operation. 
The extension is termed LL (LiDAR Loop). 
First, all required sensor data was recorded while driving. 
The evaluation was performed afterwards on a 2.60GHz Intel Core i7-4720HQ CPU with 8 GB of RAM running Linux, so that both methods had the same data available. 
Predicting with the detector takes on average 2ms and RTAB-Map's time threshold was set to 0.75s. 
Depending on the cloud sizes, registration of a loop pair takes 2s--10s. 
Due to the use of parallel threads, the longer time of registration is unproblematic.
The parameters of our extension must be adjusted depending on the environment and robot and were chosen as follows:
\begin{itemize}
\item  $r_\textrm{min}=t_\textrm{max}=3\textrm{m}\,$$\rightarrow\,$loop distance during training,
\item $n_\textrm{p,max}=10,000$, $n_\textrm{p,min}=7,000$, $n_\textrm{inliers}=1,000$, \mbox{$i_\textrm{lim}=5$,} $z_\textrm{lim}=0.3\textrm{m}$, $r_\textrm{lim}=30\textrm{m}$, $l=0.03\textrm{m}\,$$\rightarrow\,$fast and robust scan registration,
\item $n_\textrm{n,max}=200\,$$\rightarrow\,$many nodes during loop search and computation time less than RTAB-Map's time threshold,
\item $\beta=0.25\,$$\rightarrow\,$reduces the increase of the search radius,
\item $n_\textrm{v}=1\,$$\rightarrow\,$strict verification,
\item $\alpha_\textrm{min}=0.5, n_\textrm{ms}=2, r_\textrm{ms}=5\textrm{m}, n_\textrm{start}=3\,$$\rightarrow\,$global loop search until localization is satisfactory.
\end{itemize}

\begin{figure}[tbp]
	\centerline{\includegraphics[width=\linewidth]{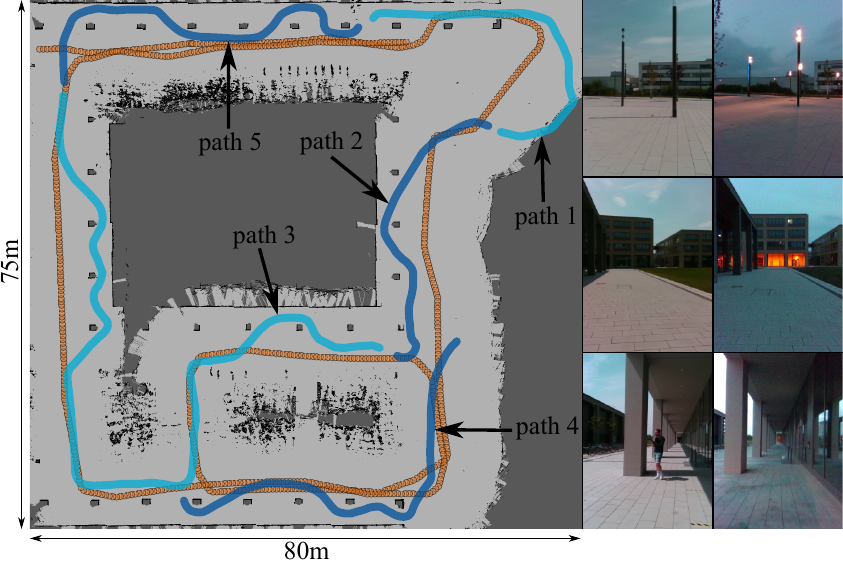}}
	\caption{Outdoor map with associated nodes (orange circles), five paths (light and dark blue) that were driven on another day and images of three exemplary loop pairs detected by RTAB-Map+LL.}
	\label{fig:outdoor_experiments}
\end{figure}

According to odometry, the path for the outdoor map has a length of 423m and consists of 769 nodes. 
When mapping with RTAB-Map, there were 6 visual loop closures, compared to 15 loops closed with the extended method. 
The potential of the extension is already apparent in single session operation, as significantly more loops were detected. 
However, there are no major differences in terms of map quality. 
The main reason is the precise odometry used, which means that with such a path, a few loops are sufficient for an acceptable map quality. 
Due to similar maps, Fig.~\ref{fig:outdoor_experiments} presents only the map generated by RTAB-Map+LL with the corresponding nodes in orange.

The added value becomes evident when taking the usual task of a mobile robot into account: 
\textit{map the environment and extend the map on another day}. 
The central requirement for such a multi-session operation is a successful localization in the old map. 
For this, the paths shown in Fig.~\ref{fig:outdoor_experiments} in light and dark blue were driven on another day at a different time. 
It was checked individually whether a localization succeeded or failed in the map generated with the respective method. 
Due to the dynamic environment, RTAB-Map did not detect any loops and, therefore, a localization for all of the five paths failed, preventing to extend the map. 
In contrast, with RTAB-Map+LL, a localization in the map of the old session was possible in four out of five paths, and the mapping could be continued. 
Loops were detected in the third path, but localization failed due to a wrong transformation matrix. 
The participating scans are from a location on the campus which has few descriptive elements except for a multitude of repetitive pillars. 
Due to the repetitiveness, pillars of one scan were registered with different pillars of the other scan, so that a translational error led to a mapping failure.

To illustrate the environmental dynamics, Fig.~\ref{fig:outdoor_experiments} shows pictures of loop pairs successfully added in RTAB-Map+LL operation. 
A change in sunlight, switched on lights, a person in the picture and a different field of view are the reasons why the visual loop search of RTAB-Map failed. 
Only by observing the three-dimensional surrounding structure, a localization is still possible. 
Even if the robot travels to an already visited location with a different orientation and therefore the image comparison returns no match, loop detection succeeds due to the 360° view and the rotation invariant features used.

Similar experiments were conducted in two challenging indoor environments: 
an entrance hall with many glass fronts and an office environment with corridors. 
According to odometry, the mapping paths have a length of 196m and 299m respectively. 
Analogous to the outdoor experiment, three paths were driven at a different time in each environment. 
RTAB-Map operation realizes successful localization in two out of six paths, whereby in the respective areas, a change in sunlight has a small influence due to a large amount of artificial light. 
Using RTAB-Map+LL, a successful localization is present in all six cases. 
Combining the results of the indoor and outdoor validation, RTAB-Map realizes a successful localization in two out of eleven cases.
In contrast to that, with the LiDAR extension it was possible to localize in ten out of eleven paths.
\subsection{KITTI Dataset}\label{kitti_sec}
Finally, our extension was evaluated with the widely used KITTI\cite{Geiger.2012b} odometry benchmark.
We continued to use the detector trained in Sec.~\ref{det_perf}, so that the training data recorded in indoor and outdoor environments on our campus differs significantly from the test data acquired in road traffic.
The sequences containing loops are mapped with our proposed extension using the RTAB-Map parameters for KITTI \cite{Labbe.2018b}.
Odometry is calculated with the front stereo camera image sequences Frame-To-Map (F2M) wise. 
The following parameters of our package have been adjusted for the car and road traffic:
$r_\textrm{min}=7.5\textrm{m}$, $r_\textrm{max}=50\textrm{m}$, $n_\textrm{v}=3$, $z_\textrm{lim}=1\textrm{m}$, \mbox{$l=0.2\textrm{m}$}, $t_\textrm{max}=10\textrm{m}$.
Since the intensities of the point clouds were not given, we set $i_\textrm{lim}=0$.

Our extended version of RTAB-Map is compared to the LiDAR-based approach LOAM \cite{zhang2017} which is currently ranked \#2 on KITTI's odometry leaderboard.
Table \ref{tab:kitti} shows the results of the two methods using the average translational error as the performance metric.
The number of visual and the number of additional LiDAR-based loop closures in RTAB-Map+LL operation are also included.
The results show that the performance of our extension is comparable to LOAM. 
In four out of seven sequences, the error is lower with our method than with LOAM. 
For sequence 09, RTAB-Map+LL performs significantly worse. 
This can be explained by the small number of loop closures.
Moreover, the potential of the LiDAR extension can be seen in sequence 08.
Since the camera is oriented in the opposite direction when traversing back, preventing the detection of visual loops.
By using LiDAR data, 30 loops can still be closed in this case.

\begin{table}[t]
	\caption{Results for the KITTI odometry dataset. Average translational error in \%.}
	\begin{tabular}{|l|rrrrrrr|}
	\hline
	Sequence&00&02&05&06&07&08&09\\
	\hline
	RTAB-Map+LL& \textbf{0.70}&0.99&\textbf{0.39}&\textbf{0.60}&\textbf{0.58}&1.18&1.39\\
	\# visual loops&156&51&88&54&18&0&3\\
	\# LiDAR loops&126&5&72&38&18&30&1\\
	\hline
	LOAM \cite{zhang2017}& 0.78&\textbf{0.92}&0.57&0.65&0.63&\textbf{1.12}&\textbf{0.77}\\
	\hline
	
\end{tabular}

	\label{tab:kitti}
\end{table}

This experiment demonstrates the general applicability of our method. 
Despite large deviation between training and test environment, loops are still detected successfully. 
To improve loop detection in road traffic, training could be carried out directly with part of the KITTI data. 
In this case, the loop distance of 3m defined during training could also be increased, since detection is desired at greater distances on the road.

\section{Conclusions}\label{conclusions}
We presented a module to close loops based on laser scans to extend graph-based SLAM methods. 
By global description of a point cloud with rotation invariant features and a trained classifier, the question \textit{Have I been here before?} can be answered under challenging conditions, which is essential for the long-term autonomy of mobile robots. 
Experiments show that the classifier detects $47.3\%$ of loops with a small number of false positives which can be filtered out by verification. 
In dynamic environments, localization with RTAB-Map succeeded only in two out of eleven cases.
A connection between all different sessions could be established using our extension. 
Except for a registration error, in ten out of eleven experiments, the localization succeeded.
The potential of our module was also demonstrated by a validation on the KITTI dataset.

Despite all the positive results, new tasks arise: 
Since registration can fail with many repeating elements, the acceptance of a calculated transformation should be further restricted, e.g. by considering the percentage of overlapping regions. 
Further, we will integrate the extension into a map management approach \cite{Ehlers.2020}. 
With different SLAM configurations, it would be possible to use a different classifier and registration parameters for a short range than for a long range environment. 
The main potential for improvement lies in the use of absolute positions using WiFi and GPS data. 
When starting multi-session operation instead of a loop search in the entire map, loops could be searched for directly in the local environment.  
\bibliographystyle{IEEEtran}
\bibliography{literatur}
\end{document}